\documentclass[letterpaper, 10 pt, conference]{ieeeconf}  

\usepackage{soul,color}
\usepackage{amsmath}
\usepackage{cancel}
\usepackage{graphicx}
\usepackage{subcaption}
\usepackage{cite}
\usepackage{booktabs}
\IEEEoverridecommandlockouts                              

\title{\LARGE \bf
Evaluating Gait Symmetry with a Smart Robotic Walker: A Novel Approach to Mobility Assessment
}

\author{Mahdi Chalaki$^{1,4}$, Abed Soleymani$^{1}$, Xingyu Li$^{1}$, Vivian Mushahwar$^{2,4}$, and Mahdi Tavakoli$^{1,3,4}$
\thanks{*This research was supported by the Canada Foundation for Innovation (CFI), the Natural Sciences and Engineering Research Council (NSERC) of Canada, the Canadian Institutes of Health Research (CIHR), Alberta Innovates, and the Government of Alberta’s grant to Centre for Autonomous Systems in Strengthening Future Communities}
\thanks{$^{1}$M. Chalaki, A. Soleymani, X. Li, and M. Tavakoli are with the
Department of Electrical and Computer Engineering, University of
Alberta, Edmonton, AB, Canada. {\tt\small mahdichalaki, zsoleymani, xingyu, mahdi.tavakoli}@ualberta.ca }%
\thanks{$^{2}$V. Mushahwar is with the Department of Medicine, Division of Physical Medicine and Rehabilitation, University of Alberta
        {\tt\small vivian.mushahwar@ualberta.ca}}%
\thanks{$^{3}$M. Tavakoli is with the Department of Biomedical Engineering, University of Alberta}%
\thanks{$^{4}$M. Chalaki, M. Tavakoli, and V. Mushahwar are with the Institute for Smart Augmentative and Restorative Technologies and Health Innovations (iSMART), University of Alberta}%
}

\begin{document}
\maketitle
\thispagestyle{empty}
\pagestyle{empty}

\begin{abstract}

Gait asymmetry, a consequence of various neurological or physical conditions such as aging and stroke, detrimentally impacts bipedal locomotion, causing biomechanical alterations, increasing the risk of falls and reducing quality of life.
Addressing this critical issue, this paper introduces a novel diagnostic method for gait symmetry analysis through the use of an assistive robotic Smart Walker equipped with an innovative asymmetry detection scheme.
This method analyzes sensor measurements capturing the interaction torque between user and walker. By applying a seasonal-trend decomposition tool, we isolate gait-specific patterns within these data, allowing for the estimation of stride durations and calculation of a symmetry index.
Through experiments involving $\boldsymbol{5}$ experimenters, we demonstrate the Smart Walker's capability in detecting and quantifying gait asymmetry by achieving an accuracy of $\boldsymbol{84.9\%}$ in identifying asymmetric cases in a controlled testing environment. Further analysis explores the classification of these asymmetries based on their underlying causes, providing valuable insights for gait assessment.
The results underscore the potential of the device as a precise, ready-to-use monitoring tool for personalized rehabilitation, facilitating targeted interventions for enhanced patient outcomes.
\end{abstract}

\section{INTRODUCTION}
\subsection{Background}

Bipedal locomotion, relying on synchronized limb movements controlled by the central nervous system \cite{viteckova_gait_2018}, plays a vital role in our independence, quality of life, and psychological well-being. It fosters self-esteem, social interaction, and facilitates independent movement across diverse environments \cite{hsiao-wecksler_review_2010,ZANIN20223257,wang_walking_2016}.
In typical adults, gait is generally symmetrical, enhancing energetic efficiency and reduces wear and tear on joints and cartilage \cite{Cabral2017}.
This bilateral symmetry, however, can be disrupted by age, disease, and injury, impacting the overall quality of mobility and walking/running functions \cite{Cabral2017}.

Increased gait asymmetry in stroke survivors \cite{allen2011step}, and persons with limb loss \cite{devan_spinal_2015} often stems from compensatory strategies to manage pain or altered mechanics \cite{Cabral2017}.
These asymmetries impact various biomechanical aspects of movement, including speed of walking, energy efficiency, and stability \cite{PATTERSON2008304}, and can contribute to reduced bone density and increased load on the contralateral limb \cite{viteckova_gait_2018}.
Assessing gait deviations may present an intuitive way for evaluating overall health, improving quality of life, and predicting outcomes such as cognitive decline and fall risk \cite{https://doi.org/10.1111/j.1468-1331.2009.02612.x}.

While asymmetry can signify an impairment, it is also present in typical individuals, albeit to a lesser degree, emphasizing its relevance in distinguishing normal from pathological gait \cite{viteckova_gait_2018}.
Modern gait assessments primarily focus on quantifying asymmetries in individuals with gait disorders, monitoring post-injury or surgical recovery, and examining asymmetries in healthy gait concerning lateral dominance \cite{viteckova_gait_2018}.
Research also delves into refining gait asymmetry measurements to detect the stage of diseases such as Parkinson's \cite{Santanna20112127},
suggesting that lower asymmetry is observed in the early stages of the disease and gradually increases with disease progression.
This offers a promising avenue for diagnosis and treatment tailoring.
Beyond diagnosis, the focus can shift to rehabilitation and restoring symmetry.
Techniques like split-belt treadmill training for stroke survivors and real-time feedback programs for hip replacement patients demonstrate success in improving step length, spatiotemporal, and loading symmetry \cite{WHITE20051958}.

\subsection{Challenges in Gait Assessment: A Need for Innovation}
Gait analysis typically occurs in controlled environments  with specialized tools, including pressure sensors in walkways \cite{WEBSTER2005317}, and optical motion capture systems such as Time-of-Flight Systems (ToF) \cite{abellanas_ultrasonic_2008}.
However, these methods are confined to specific areas and require expert setup and control, which can be distressing for individuals \cite{ballesteros_gait_2016}.
Alternatively, gait analysis can be conducted using wearable sensors, such as inertial sensors on legs \cite{10.1007/978-3-642-34546-3_152}, and electromyogram (EMG) electrodes on muscles of interest \cite{WENTINK2014391}. However, these solutions may be uncomfortable for users to wear for extended periods, and they also present other complications such as noise and cross talk in EMG activity \cite{mesin_crosstalk_2020}, and integration drift problems in inertial sensors \cite{NAZARAHARI2022110880}. Some of these methods, while valuable, lack real-world practicality, and restrict early detection and intervention for gait disorders.

 \begin{figure*}[t!]
    \centering
	\subfloat{\includegraphics[width=0.85\textwidth]{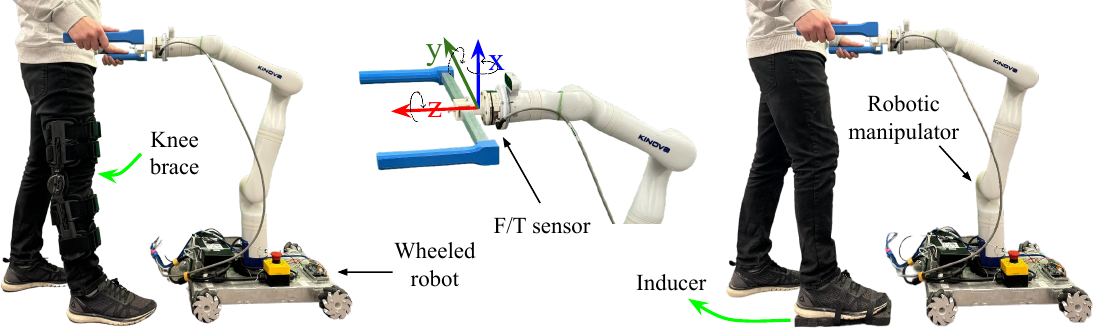}}
	\caption{(\textit{left}) Illustration of mobility assistance with the Smart Walker, knee immobilizer induces hemiplegic gait on the right leg - (\textit{middle}) Coordinate
reference frame on the force/torque sensor - (\textit{right}) Demonstration of simulating leg length discrepancy to mimic short-leg gait.}
	\label{fig:setup}
\vspace{-15pt}
\end{figure*}

\subsection{Instrumented rollators: A Promising Solution}
To enhance usability and facilitate gait analysis in daily life, sensors can be integrated into commonly used devices, like smartphones or assistive devices such as walkers and rollators \cite{ballesteros_online_2016}.
Rollators equipped with force sensors can provide continuous data on gait parameters by detecting changes in weight distribution on handlebars, indirectly indicating heel strike \cite{ballesteros_gait_2016}.
While instrumented rollators have been used for gait analysis with optical sensors like Microsoft Kinect or ToF cameras \cite{abellanas_ultrasonic_2008}, these vision-based solutions face limitations regarding illumination control, setup complexity, and data processing requirements.

The present study introduces a unique method based on the analysis of measurements obtained from rollator-embedded force sensors.
This method offers a less intrusive and more continuous way to assess and monitor gait symmetry in real-world settings compared to conventional gait analysis methods that require specialized equipment and environments.
The integration of force-based gait analysis into rollators provides several advantages: affordability, reliability, and user stability \cite{YOUDAS2005394}.
Additionally, this approach can be seamlessly integrated into daily life, offering valuable evaluation of gait regardless of location or time constraints.
This makes the technology suitable for use in both clinical settings, providing objective data to healthcare professionals, and at home, empowering individuals to monitor their gait patterns over time.
Furthermore, the potential of the Smart Walker's robotic arm can be explored for functionalities beyond gait analysis, such as assisting with sit-to-stand maneuvers and carrying personal belongings, enhancing its overall utility.

Wang \textit{et al.} \cite{wang_walking_2016} utilized a cost-effective rollator system with odometry and a gyroscope and found significant group differences in accuracy, ability, and stability, suggesting its potential for age-related gait assessment.
Ballesteros \textit{et al.} \cite{ballesteros_online_2016} proposed an unsupervised method to predict the Tinetti Mobility test score using on-the-fly spatiotemporal gait parameters from a smart rollator.
By correlating these parameters with the Tinetti test, their study aims to identify key indicators that can assess an individual's condition without requiring the Tinetti test itself.
Ojeda \textit{et al.} \cite{ojeda_automatic_2018} developed an unsupervised method using the i-Walker to categorize older individuals' walking patterns into four groups based on gait characteristics, distinguishing between young individuals based on speed and identifying geriatric gait.

\subsection{Research Objectives and Significance}
Our work focuses specifically on gait temporal symmetry.
We propose a method that leverages the capabilities of rollator-embedded force sensors to achieve three key objectives:
\begin{enumerate}
    \item Detect asymmetry patterns: We aim to identify a person's gait based on their symmetry patterns, potentially aiding in the early detection of gait abnormalities.
    \item Measure the severity of asymmetry: We will quantify the degree of asymmetry within each individual's gait pattern, providing a numerical measure of their gait deviation from symmetry.
    \item Gait asymmetry classification: To explore further, we aim not only to detect asymmetry but also to categorize the type of gait disorder responsible for it. This provides clinicians with essential information for interventions and rehabilitation strategies.
\end{enumerate}
To the best of our knowledge, no prior study has utilized force estimation on rollators for gait symmetry analysis. By addressing this gap in research, our work holds the potential to enhance the diagnostic capabilities of rollator-based gait analysis, provide objective and continuous monitoring of gait symmetry in real-world settings, and contribute to the development of personalized interventions and rehabilitation strategies for individuals with gait impairments.

The paper is structured as follows: Section II details a mathematical approach to isolating gait-related components and introduces the wheeled mobile manipulator (WMM) utilized as a Smart Walker. Section III presents experimental results with the Smart Walker device, followed by a discussion of the results. Section IV concludes the research and outlines future work.

\begin{figure*}[t!]
    \centering
	\subfloat{\includegraphics[width=0.85\textwidth]{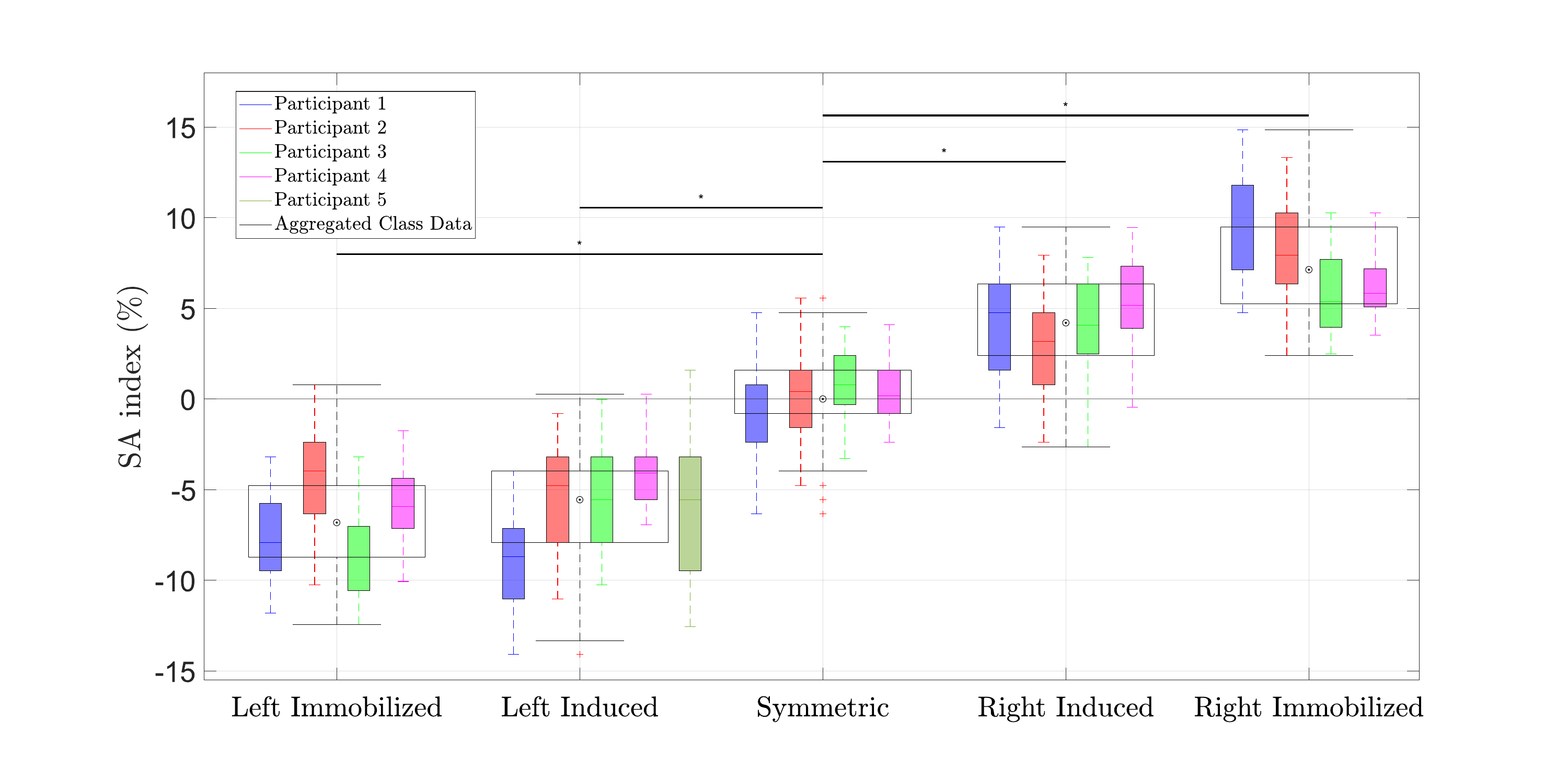}}
	\caption{Distribution of symmetry angles among experimenters across five gait classes.}
	\label{fig:Classes_211}
\vspace{-15pt}
\end{figure*}

\section{Methodology} \label{Methodology}

The human trunk and center of mass exhibit cyclical displacements in the sagittal plane during gait \cite{MOSTAFAVI2022102173}.
In walker-assisted gait, similar cyclical patterns reflect in the interaction forces between the user and the walker handles \cite{sierra_m_humanrobotenvironment_2019}, allowing for the extraction of spatiotemporal variables such as step length, step time, and stance or swing phase duration.
For instance, Alwan \textit{et al.} \cite{alwan_basic_2007} demonstrated a correlation between forces and moments in the vertical direction of walker handles and heel marker trajectories, with force and moment peaks aligning with heel initial contact.This highlights the potential for force sensors to capture key gait events. The predominant approach to assessing symmetry involves using the extracted features to compute a single discrete index, summarizing symmetry across spatiotemporal, kinetic, or kinematic parameters \cite{viteckova_gait_2018}.
However, utilizing interaction force or torque presents specific challenges.
Individuals may distribute weight between their legs and the walker differently, or exert varying force levels on the handles.
Additionally, muscle fatigue can vary significantly day-to-day, impacting force patterns and potentially influencing the gait monitoring process.
Further limitations include user tiredness during extended walking tests and potential changes in walking speed during trials.
To effectively address these challenges and ensure robust gait analysis, we proposed a novel methodology that is resilient to variations in interaction force magnitude, user cadence, test duration, and muscle fatigue levels.
This approach is detailed in the next section.

\subsection{Seasonal-Trend Decomposition of Interaction Force}
We employed a comprehensive approach to uncover the asymmetrical behaviors in human walking data.
Specifically, by applying the seasonal-trend decomposition method \cite{additive_stl} to the recorded time series data of the user interaction torque, we gained valuable insights into the temporal and spatial characteristics within the given trajectory, revealing subtle variations and anomalies \cite{soleymani2022surgical}.
The primary reason to choose such a method for analyzing human walking data originates from the quasi-periodic nature of human walking, which requires a methodological framework capable of separating the inherent periodicity from other sources of variability.
A \textit{season} in this framework refers to repetitive patterns or cycles in walking behavior that occur at regular intervals, combined with modest periodic changes related to physical or environmental factors.
On the other hand, a \textit{trend} denotes a long-term progression or shift in a trial of walking data, indicating gradual changes that could be due to user fatigue during extended testing, or variations in walking cadence.
The decomposition of the walking time series into general trends, seasonal patterns, and residuals facilitates the extraction of meaningful information at different temporal channels, enabling a multi-resolution analysis of walking dynamics.
This hierarchical approach enhances interpretability and provides a deeper insight into the complex interplay between various factors influencing human gait.

In this way, capturing long-term trends and systemic deviations from the expected flow of movement provides insights into global aspects of gait dynamics, including factors such as fatigue, muscle weakness, and biomechanical inefficiencies.
On the other hand, asymmetry in walking, manifests as deviations from the expected periodic patterns.
In this way, seasonal patterns highlight and quantify the partial abnormal behaviors linked specifically to asymmetrical walking.
The residual signal, resulting from the decomposition, contains the data that need to be filtered out, which primarily includes sensor noise and walking effects irrelevant to the asymmetrical behaviors being studied.
By discarding these unnecessary elements, we focused our analysis exclusively on the aspects of walking associated with asymmetry, enhancing the precision and relevance of our findings.

In seasonal-trend decomposition, two primary methods are commonly utilized: additive and multiplicative.
The additive method is better suited for our application as it assumes that the seasonal fluctuations remain constant across different levels of the trend.
The additive model for time series decomposition attempts to reconstruct the observed time-series ${\cal X}(t)$ based on an estimation $\hat{{\cal X}}(t)$ as follows:
\begin{equation}
{\cal X}(t) \approx \hat{{\cal X}}(t) = {\cal T}(t) + {\cal S}(t) + {\cal R}(t),
\label{eq:additive}
\end{equation}
where ${\cal T}(t)$, ${\cal S}(t)$, and ${\cal R}(t)$ represent the trend component, seasonal component, and residual (or error) component, respectively, at time-stamp $t$.
While the additive method outlined in \eqref{eq:additive} is a common and effective approach for time series decomposition \cite{additive_stl}, its assumption of a single, fixed seasonal signal ($S(t)$) across different strides might oversimplify the intricate dynamics of human walking in our specific context.
Under fixed environmental conditions, human walking is a complex motor activity influenced by various individual factors, such as inherent structural imbalances or differences in left and right ergonomic configurations, leading to potential deviations in the periodical patterns of each stride.
\begin{figure}[t!]
    \centering
	\subfloat{\includegraphics[width=0.48\textwidth]{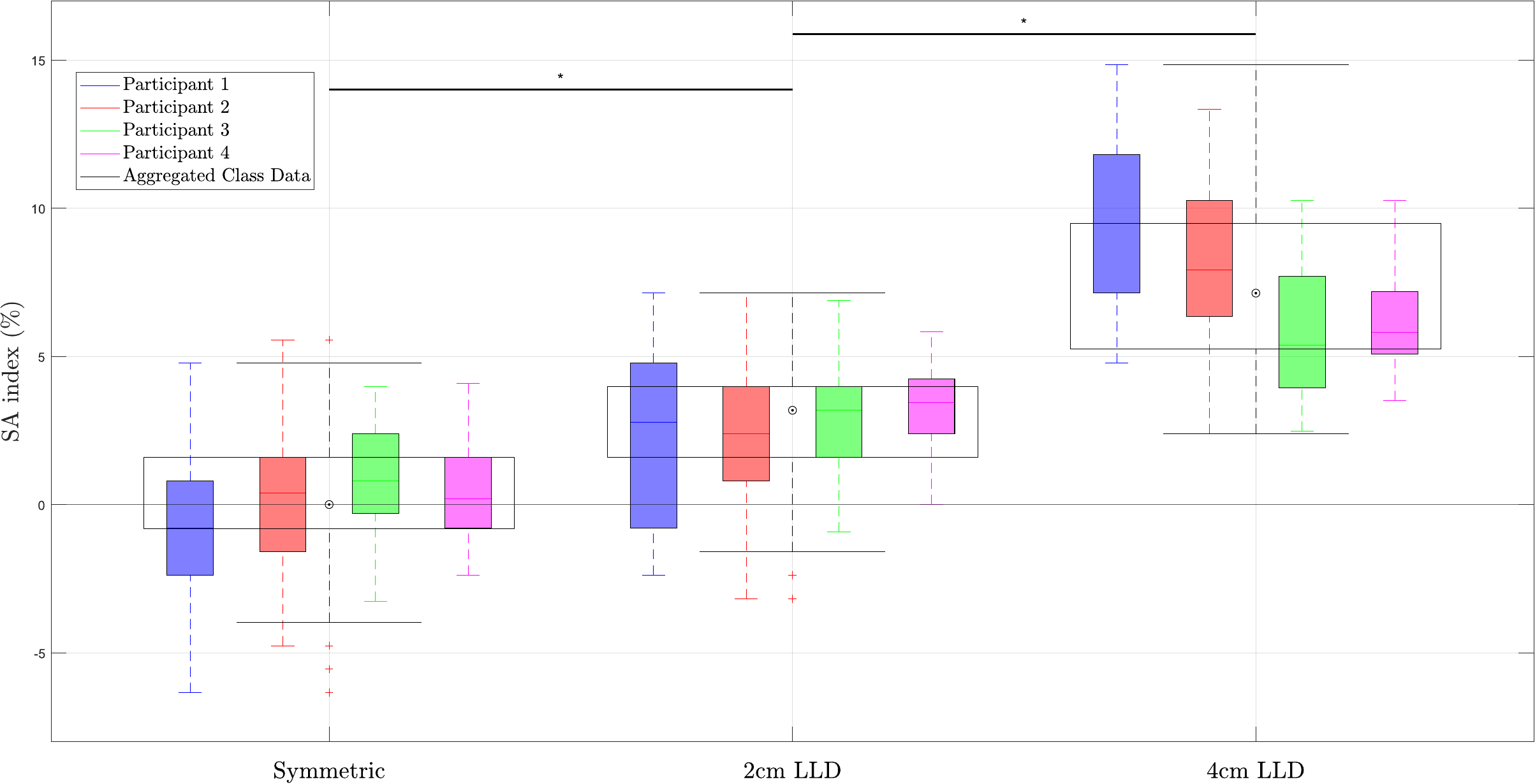}}
	\caption{The differences in symmetry angles among experimenters across two levels of LLD.}
	\label{fig:LLD}
\vspace{-15pt}
\end{figure}
To address such a challenge, a more advanced approach, particularly Multiple Seasonal-Trend decomposition using Locally estimated scatterplot smoothing (MSTL) \cite{bandara_mstl_2022}, was chosen for its ability to disentangle the quasi-periodic nature of human walking.
In this way, \eqref{eq:additive} is to be reformulated as follows:
\begin{equation}
\hat{{\cal X}}(t) = {\cal T}(t) + \sum_{i=1}^{n} w_i\times S_i(t) + {\cal R}(t),
\label{eq:MSTL}
\end{equation}
where $w_i$ highlights the contribution of each subseasonality $S_i(t)$ in total seasonality ${\cal S}(t)$, and $n$ denotes the intrinsic number of seasonalities within the time-series. In the context of gait analysis, each of these periodicities corresponds to a single stride, and the method used to calculate $n$ is elaborated upon in the following section.
Assuming that the residuals, denoted as ${\cal R}(t)$, are drawn from a Gaussian distribution, we employed root mean square error (RMSE) as the optimization metric and the calculation of decompositions in \eqref{eq:MSTL}.

\subsection{Experimental Setup}
Our experimental setup leveraged a versatile wheeled mobile manipulator (WMM) transformed into a smart walker to facilitate gait analysis. The system consists of a mobile base, a custom-built four-wheel platform equipped with mecanum wheels for omnidirectional mobility; a robotic arm, specifically a $7$-DOF Kinova Gen3 arm (Kinova Robotics, Canada), which provides flexibility for height adjustment and potential for future assistance; and a force/torque sensor, an Axia$80$-ZC$22$ F/T sensor (ATI Industrial Automation, Apex,
NC, USA) integrated into the handles, capturing user interaction forces during gait analysis experiments (please refer to Fig. \ref{fig:setup} for more details).

\subsection{Data Collection}

While healthy gait can be disrupted by various disorders, this study focuses on two distinct conditions that significantly impact walking symmetry: hemiplegic gait and short leg gait. Hemiplegic gait, often a consequence of stroke or other neurological conditions, affects a considerable portion of the global population. It leads to reduced mobility, increased fall risk, and characteristic movement patterns involving reduced stride length and knee flexion, causing one leg to appear stiff and swing out during walking\cite{STEIN2020931}.
Short leg gait, on the other hand, arises from congenital factors or injuries. It presents challenges in daily activities and can lead to long-term musculoskeletal issues if not properly addressed. This condition, also known as Leg Length Discrepancy (LLD), is characterized by a noticeable difference in leg lengths, impacting stability during standing and walking. LLD typically results from changes in lower limb kinematics, including pelvic tilt, and hip and knee flexion \cite{STEIN2020931}.
To analyze the effect of these impairments on gait symmetry in a controlled setting, we simulated them on each leg. Here, the Left and Right Leg Induced (LI and RI) datasets represent simulations of LLD achieved by artificially increasing the effective length of the a leg during data collection (e.g., attaching a 4 cm object under the shoe). Conversely, the Left and Right Leg Straight (LS and RS) datasets simulate hemiplegic gait by using a commercial knee immobilizer (ZJchao, China) to restrict knee flexion, mimicking the characteristic stiffness. This approach allows us to investigate how each specific impairment disrupts gait symmetry.

This study used the $10$-Meter Walk Test, a well-established and straightforward method for evaluating mobility. Experimenters walked a straight 10-meter path with the Smart Walker, and each of the five data classes underwent ten trials per test subject. To ensure repeatability and reliable statistical analysis, trials were conducted on consecutive days, with one trial per class each day to minimize muscle fatigue. Data analysis focused on the middle 10 meters of a 14-meter walkway, following the standard protocol to exclude initial acceleration and final deceleration phases, ensuring only steady-state gait was analyzed.

\subsection{Data Analysis}

During each $10$-meter walking trial, the torque on the walker handles was recorded at $40$ Hz using the ATI sensor. To remove high-frequency noise arising from walker-ground vibrations, the recorded torque signal was then filtered using a $4^{\text{th}}$-order Butterworth low-pass filter with a cut-off frequency of $5$ Hz.
To align with the seasonal-trend decomposition method, the number of seasons (equivalent to strides) in a walking trial was determined using peak detection techniques, and the length of each season was standardized by upsampling. The seasonal pattern was then extracted by applying MSTL to the filtered signal. To facilitate the comparison of left and right step characteristics and evaluate gait symmetry, we utilized stride-level information, encompassing both steps within a single unit. Consequently, we segmented the seasonal pattern into strides.
By identifying inflection points in each segment, we can accurately detect step occurrences within each stride \cite{ballesteros_gait_2015}. Due to the normalized stride durations, we expressed step duration as a percentage of the entire stride, denoted as $ t_{1} $ and $ t_{2} $ for the left and right foot, respectively.
In the literature, single symmetry measures are used to compare symmetry levels across groups or following interventions \cite{viteckova_gait_2018}.
This study used the symmetry angle (SA) index \eqref{eq:SA}, which has demonstrated superior robustness compared to the symmetry index (SI) regarding issues related to reference value and inflation \cite{zifchock_symmetry_2008}, and is formulated as follows: 
\begin{equation}
\mathrm{SA}=\left(45^{\circ}-\arctan \left(\frac{X_{\text {left}}}{X_{\text {right}}}\right)\right) \times \frac{100 \%}{90^{\circ}}.
\label{eq:SA}
\end{equation}
Equation \eqref{eq:SA} assesses symmetry by evaluating the angle formed between discrete values gathered from the left and right sides ($ X_\text{left} $, $ X_\text{right} $). Using step duration measurements for the left and right foot, $ t_{1} $ and $ t_{2} $ were substituted into the equation to compute the SA index, where $0\%$ indicates perfect symmetry and $100\%$ signifies opposite equal values.
Following SA index computation for every stride in each trial, paired $t$-tests were conducted to assess statistically significant differences between the symmetric class and each asymmetric class. This step assessed the efficacy of our proposed method in detecting gait pattern asymmetry.
Beyond detecting gait asymmetry, this study aimed to evaluate the smart walker's potential as a longitudinal gait monitoring tool. To this end, an experiment was conducted involving two objects of varying heights to simulate different levels of leg height discrepancy. The results from four of the experimenters with symmetric gait were analyzed to evaluate the method's ability to detect the extent of asymmetry.

\begin{table}[t!]
    \centering
    \caption{Statistical analysis of users' gaits across five classes with $95\%$ Confidence Intervals.}
    \label{tab:comprehensiveTable}
    \resizebox{0.48\textwidth}{!}{
    \begin{tabular}{cccccc}
        \toprule
        Class & Experimenter & CI $(\sigma, \Sigma)$ & Mean & SEM  & $p$-value \\
        \midrule\midrule
        & $1$ & $(3.77, 4.76)$ & $4.26$ & $0.25$ & $<0.0001$ \\
        & $2$ & $(2.26, 3.17)$ & $2.72$ & $0.23$ & $<0.0001$ \\
        RS & $3$ & $(3.71, 4.47)$ & $4.09$ & $0.19$ & $<0.0001$ \\
        & $4$ & $(4.95, 5.67)$ & $5.31$ & $0.18$ & $<0.0001$ \\
        \midrule
        & $1$ & $(-8.15, -7.29)$ & $-7.72$ & $0.22$ & $<0.0001$ \\
        & $2$ & $(-4.96, -4.02)$ & $-4.49$ & $0.24$ & $<0.0001$ \\
        LS & $3$ & $(-8.96, -8.23)$ & $-8.59$ & $0.18$ & $<0.0001$ \\
        & $4$ & $(-6.04, -5.36)$ & $-5.70$ & $0.17$ & $<0.0001$ \\
        \midrule
        & $1$ & $(-1.49, -0.7)$ & $-1.09$ & $0.20$ & -- \\
        & $2$ & $(-0.53, 0.64)$ & $0.06$ & $0.29$ & -- \\
        Sym & $3$ & $(0.72, 1.28)$ & $1.00$ & $0.14$ & -- \\
        & $4$ & $(0.1, 0.62)$ & $0.36$ & $0.13$ & -- \\
        \midrule
        & $1$ & $(8.95, 9.87)$ & $9.41$ & $0.23$ & $<0.0001$ \\
        & $2$ & $(7.64, 8.64)$ & $8.14$ & $0.25$ & $<0.0001$ \\
        RI & $3$ & $(5.45, 6.22)$ & $5.83$ & $0.19$ & $<0.0001$ \\
        & $4$ & $(6.02, 6.6)$ & $6.31$ & $0.15$ & $<0.0001$ \\
        \midrule
        & $1$ & $(-9.49, -8.53)$ & $-9.01$ & $0.24$ & $<0.0001$ \\
        & $2$ & $(-5.75, -4.78)$ & $-5.27$ & $0.24$ & $<0.0001$ \\
        LI & $3$ & $(-5.99, -5.14)$ & $-5.56$ & $0.22$ & $<0.0001$ \\
        & $4$ & $(-4.56, -4.13)$ & $-4.34$ & $0.11$ & $<0.0001$ \\
        \bottomrule
    \end{tabular}}
\vspace{-15pt}
\end{table}

\section{EXPERIMENTS AND RESULTS}

\subsection{Experiment 1: Simulating Gait Disorders for Symmetry Assessment}

In Experiment 1, five experimenters participated: three men and two women. Four had typical gait, while one had pre-existing asymmetric gait. The four participants with typical gait completed tasks representing four simulated classes in addition to their normal gait, resulting in a dataset with five classes. The participant with asymmetric gait completed 10 trials of the experiment without any attachments.

Following the data analysis part, the single numerical SA index was computed for each stride. Quantile-quantile plots were employed to evaluate the conformity of the data distribution within each class to a theoretical normal distribution, a prerequisite for employing the $t$-test.

\subsubsection{Evaluation of the Smart Walker's Discriminatory Capacity}To gauge the proficiency of the smart walker in differentiating symmetric gait from various forms of asymmetric gait, we employed the SA index computed from individual strides. The distribution of experimenter data is visualized using box plots in Fig.~\ref{fig:Classes_211}.
We conducted paired $t$-tests to compare data recorded for each individual across different gait classes on the same days, considering a significance level of $p < 0.001$. The null hypothesis assumed equality of means between the populations ($\mu_1 = \mu_2$).
Table \ref{tab:comprehensiveTable} presents the lower ($\sigma$) and upper ($\Sigma$) bounds of the Confidence Interval (CI), where the mean and standard error of mean (SEM) have been used to calculate $p$-values for each comparison between symmetric and asymmetric classes. Notably, all metrics within the LLD-induced and knee-immobilized groups exhibited significant asymmetry compared to typically-walking individuals. Furthermore, the distributions illustrated variations in the direction of asymmetry across classes, including prolonged time spent on either the simulated paretic or non-paretic limb.
SA values less than zero signify a bias towards the left limb, a clinically relevant aspect potentially informative for future rehabilitation strategies.

\begin{table}[t!]
    \centering
    \caption{Statistical analysis of users' gaits across three LLD levels with $95\%$ Confidence Intervals.}
    \label{tab:LLD}
    \resizebox{0.48\textwidth}{!}{
    \begin{tabular}{cccccc}
        \toprule
        Class & Experimenters & CI $(\sigma, \Sigma)$ & Mean & SEM  & $p$-value \\
        \midrule\midrule
        & $1$ & $(-1.49, -0.7)$ & $-1.09$ & $0.20$ & $<0.0001$ \\
        & $2$ & $(-0.53, 0.64)$ & $0.06$ & $0.29$ & $<0.0001$ \\
        Sym & $3$ & $(0.72, 1.28)$ & $1.00$ & $0.14$ & $<0.0001$ \\
        & $4$ & $(0.1, 0.62)$ & $0.36$ & $0.13$ & $<0.0001$ \\
        \midrule
        & $1$ & $(1.48, 3.25)$ & $2.36$ & $0.44$ & -- \\
        & $2$ & $(1.82, 2.62)$ & $2.22$ & $0.20$ & -- \\
        2cm LLD & $3$ & $(2.62, 3.2)$ & $2.91$ & $0.15$ & -- \\
        & $4$ & $(3.07, 3.55)$ & $3.31$ & $0.12$ & -- \\
        \midrule
        & $1$ & $(8.95, 9.87)$ & $9.41$ & $0.23$ & $<0.0001$ \\
        & $2$ & $(7.64, 8.64)$ & $8.14$ & $0.25$ & $<0.0001$ \\
        4cm LLD & $3$ & $(5.45, 6.22)$ & $5.83$ & $0.19$ & $<0.0001$ \\
        & $4$ & $(6.02, 6.6)$ & $6.31$ & $0.15$ & $<0.0001$ \\
        \bottomrule
    \end{tabular}}
\vspace{-15pt}
\end{table}

\subsubsection{Assessment of Anomaly Detection Accuracy in Gait Symmetry}We evaluated the effectiveness of our proposed system in detecting gait asymmetry using established evaluation metrics: accuracy, precision, recall, and F1 score. To establish normative symmetry values, the $95\%$ CI was determined based on the SA indices of typical gaits. Any stride with an SA index falling outside of this CI was classified as asymmetric. The results indicated a precision of $0.995$, a recall of $0.815$, an F1 score of $0.896$, and an overall accuracy of $0.848$. These metrics highlight the robustness of the method in identifying anomalies in gait symmetry.

\subsection{Experiment 2: Monitoring Gait Asymmetry Progression}

This experiment investigated the Smart Walker's ability to monitor gait asymmetry over time. To simulate different levels of LLD, objects of varying heights ($2$ and $4$ cm) were placed under the right foot of the experimenters.

Ten trials were conducted for each subject across different days. Data processing followed the previously described methodology, and the class distribution is shown in Fig.~\ref{fig:LLD}. Paired $t$-tests were used to assess the Smart Walker's ability to detect changes in asymmetry levels. The results showed a statistically significant difference ($p<0.001$) between the 2 cm LLD class and both the symmetric gait and 4 cm LLD groups. This indicates that the Smart Walker can detect and quantify varying levels of asymmetry, which could be valuable for long-term monitoring in clinical settings, aiding early detection of gait deviations and personalized rehabilitation strategies.

\begin{figure}[t!]
    \centering
    \subfloat{\includegraphics[width=0.45\textwidth]{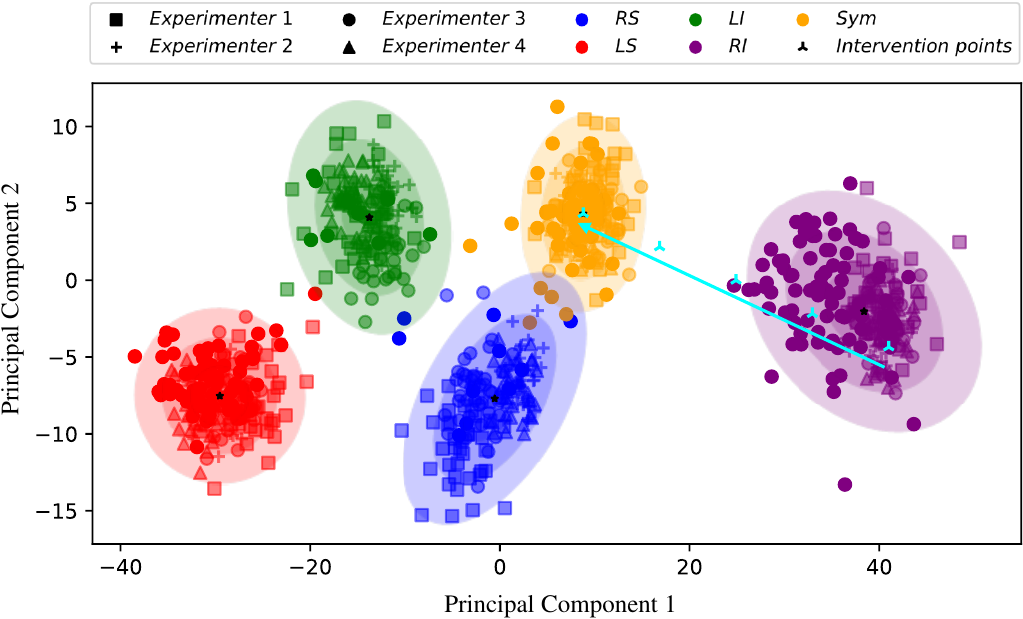}}
	\caption{GMM clustering in PCA-reduced space for the investigated five gait patterns.}
	\label{fig:GMM}
\vspace{-15pt}
\end{figure}
\subsection{Experiment 3: Gait Asymmetry Clustering}
Unsupervised learning methods, such as dimensionality reduction and clustering, have proven to be powerful tools for extracting explainable and interpretable insights from complex datasets \cite{soleymani2022domain,soleymani2023artificial}.
These techniques can reveal valuable, label-free information that often remains hidden in high-dimensional spaces.
In our study, we applied Principal Component Analysis (PCA) to the high-dimensional time series data of the seasonal component for each stride.
As illustrated in Fig. \ref{fig:GMM}, the application of PCA revealed the true nature of our dataset by identifying distinct clusters, each representing a unique walking pattern.
This finding affirms the effectiveness of our methodology in capturing the underlying features that distinguishes the symmetrical and asymmetrical behaviors exhibited during gait, all derived from the simple interaction between the robot and human.

To structurally define and mathematically formulate these clusters, we utilized the Gaussian Mixture Model (GMM); a simple probabilistic model that assumes each cluster is generated from a mixture of several Gaussian distributions.
By fitting a GMM with five components to our PCA-reduced data, we were able to mathematically define each cluster in terms of its statistical properties—such as mean and covariance.
Fig.~\ref{fig:GMM} also presents the clustering results in the reduced PCA space.
The ellipsoids encapsulate the mean and covariance of the data points, providing a structured and mathematically rigorous definition of each gait cluster.

\begin{figure}[t!]
    \centering
	\subfloat{\includegraphics[width=0.48\textwidth]{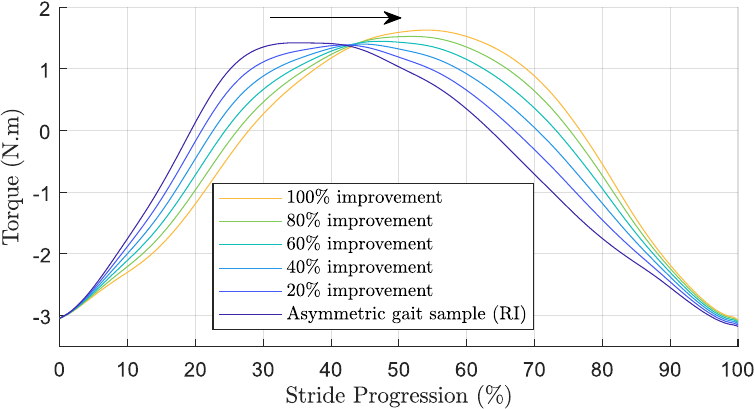}}
	\caption{Stages of gait improvement from an initially asymmetrical stride towards a symmetrical gait pattern.}
	\label{fig:Intervention}
\vspace{-15pt}
\end{figure}

Such formulation not only enhances our understanding of the separation between different gait patterns but also serves as a guide for potential future interventions.
By selecting a point from an asymmetric cluster—such as RI—we can linearly interpolate this point towards the centroid of the symmetric gait cluster, creating a virtual point \( p^* \) in PCA space. This point represents a potential improved version of the RI gait pattern.
One of the significant advantages of PCA is its preservation of the linear relationships between variables while mapping the high-dimensional data into the low-dimensional map.
Such inherent linearity in PCA allows us to posit that if \( p^* \) is incrementally moved toward the center of the symmetric cluster, the corresponding reconstructed trajectory \( \tau^* \) would reflect an improved gait pattern with a proportionate degree of correction.
Fig.~\ref{fig:Intervention} illustrates the hypothetical progressive improvements from a pure asymmetric RI gait towards a symmetric pattern.
It is observed that with each improvement step, the peak torque shifts towards the right.
This observation is consistent with our previous discussions on RI and substantiates the effectiveness of our proposed methodology for gait enhancement.
The enhanced trajectory $\tau^*$ can serve as a reference for future rehabilitative interventions aimed at correcting gait asymmetry, although its implementation is beyond the scope of this paper.
\section{CONCLUSIONS AND FUTURE WORK}
This research introduced a novel use of a wheeled mobile manipulator, configured as a Smart Walker, for analyzing gait symmetry. By utilizing an integrated torque sensor, the method extracts isolated gait patterns and compares step duration, enabling precise determination of both the direction and severity of asymmetry. The experiments demonstrated the Smart Walker's potential for enhanced diagnostic capabilities, offering continuous and objective monitoring of gait symmetry.
Looking forward, the analysis of asymmetry's extent and direction, in conjunction with the torque from human-robot interaction, sets the stage for deeper biomechanical analyses aimed at estimating the trajectory of each foot. This approach facilitates the development of strategies to correct gait asymmetries using external rehabilitation devices, such as exoskeletons. Furthermore, by integrating a framework that can detect and gather data from new, unseen classes and subsequently include them into our dataset, \cite{9729350} we envision expanding the Smart Walker's capabilities to adapt to novel gait patterns over time.
This promising avenue of research opens up possibilities for more precise, tailored intervention strategies, significantly impacting the field of rehabilitation technology.



\bibliographystyle{ieeetr}
\bibliography{Papers1}

\begin{thebibliography}{10}

\bibitem{viteckova_gait_2018}
S.~Viteckova, P.~Kutilek, Z.~Svoboda, R.~Krupicka, J.~Kauler, and Z.~Szabo, ``Gait symmetry measures: {A} review of current and prospective methods,'' {\em Biomedical Signal Processing and Control}, vol.~42, pp.~89--100, Apr. 2018.

\bibitem{hsiao-wecksler_review_2010}
E.~T. Hsiao-Wecksler, J.~D. Polk, K.~S. Rosengren, J.~J. Sosnoff, and S.~Hong, ``A {Review} of {New} {Analytic} {Techniques} for {Quantifying} {Symmetry} in {Locomotion},'' {\em Symmetry}, vol.~2, pp.~1135--1155, June 2010.
\newblock Number: 2 Publisher: Molecular Diversity Preservation International.

\bibitem{ZANIN20223257}
M.~Zanin, F.~Olivares, I.~Pulido-Valdeolivas, E.~Rausell, and D.~Gomez-Andres, ``Gait analysis under the lens of statistical physics,'' {\em Computational and Structural Biotechnology Journal}, vol.~20, pp.~3257--3267, 2022.

\bibitem{wang_walking_2016}
T.~Wang, J.-P. Merlet, G.~Sacco, P.~Robert, J.-M. Turpin, B.~Teboul, A.~Marteu, and O.~Guerin, ``Walking analysis of young and elderly people by using an intelligent walker {ANG},'' {\em Robotics and Autonomous Systems}, vol.~75, pp.~96--106, Jan. 2016.

\bibitem{Cabral2017}
S.~Cabral, {\em Gait Symmetry Measures and their Relevance to Gait Retraining}, pp.~1--19.
\newblock Cham: Springer International Publishing, 2017.

\bibitem{allen2011step}
J.~L. Allen, S.~A. Kautz, and R.~R. Neptune, ``Step length asymmetry is representative of compensatory mechanisms used in post-stroke hemiparetic walking,'' {\em Gait \& posture}, vol.~33, no.~4, pp.~538--543, 2011.

\bibitem{devan_spinal_2015}
H.~Devan, A.~Carman, P.~Hendrick, L.~Hale, and D.~C. Ribeiro, ``Spinal, pelvic, and hip movement asymmetries in people with lower-limb amputation: {Systematic} review,'' {\em Journal of Rehabilitation Research and Development}, vol.~52, no.~1, pp.~1--19, 2015.

\bibitem{PATTERSON2008304}
K.~K. Patterson, I.~Parafianowicz, C.~J. Danells, V.~Closson, M.~C. Verrier, W.~R. Staines, S.~E. Black, and W.~E. McIlroy, ``Gait asymmetry in community-ambulating stroke survivors,'' {\em Archives of Physical Medicine and Rehabilitation}, vol.~89, no.~2, pp.~304--310, 2008.

\bibitem{https://doi.org/10.1111/j.1468-1331.2009.02612.x}
O.~Beauchet, C.~Annweiler, V.~Dubost, G.~Allali, R.~W. Kressig, S.~Bridenbaugh, G.~Berrut, F.~Assal, and F.~R. Herrmann, ``Stops walking when talking: a predictor of falls in older adults?,'' {\em European Journal of Neurology}, vol.~16, no.~7, pp.~786--795, 2009.

\bibitem{Santanna20112127}
A.~Santanna, A.~Salarian, and N.~Wickstrom, ``A new measure of movement symmetry in early parkinsons disease patients using symbolic processing of inertial sensor data,'' {\em IEEE Transactions on Biomedical Engineering}, vol.~58, no.~7, p.~2127 – 2135, 2011.
\newblock Cited by: 71; All Open Access, Green Open Access.

\bibitem{WHITE20051958}
S.~C. White and R.~M. Lifeso, ``Altering asymmetric limb loading after hip arthroplasty using real-time dynamic feedback when walking,'' {\em Archives of Physical Medicine and Rehabilitation}, vol.~86, no.~10, pp.~1958--1963, 2005.

\bibitem{WEBSTER2005317}
K.~E. Webster, J.~E. Wittwer, and J.~A. Feller, ``Validity of the gaitrite® walkway system for the measurement of averaged and individual step parameters of gait,'' {\em Gait \& Posture}, vol.~22, no.~4, pp.~317--321, 2005.

\bibitem{abellanas_ultrasonic_2008}
A.~Abellanas, A.~Frizera~Neto, R.~Ceres~Ruiz, R.~Raya, and L.~Calderón~Estévez, {\em Ultrasonic Time of Flight Estimation in Assistive Mobility: Improvement of the model-echo fitting}.
\newblock VDI/VDE, Sept. 2008.
\newblock Accepted: 2009-06-15T09:05:15Z.

\bibitem{ballesteros_gait_2016}
J.~Ballesteros, C.~Urdiales, A.~B. Martinez, and J.~H. Van~Dieën, ``On {Gait} {Analysis} {Estimation} {Errors} {Using} {Force} {Sensors} on a {Smart} {Rollator},'' {\em Sensors}, vol.~16, p.~1896, Nov. 2016.
\newblock Number: 11 Publisher: Multidisciplinary Digital Publishing Institute.

\bibitem{10.1007/978-3-642-34546-3_152}
A.~Ferrari, L.~Rocchi, J.~van~den Noort, and J.~Harlaar, ``Toward the use of wearable inertial sensors to train gait in subjects with movement disorders,'' in {\em Converging Clinical and Engineering Research on Neurorehabilitation} (J.~L. Pons, D.~Torricelli, and M.~Pajaro, eds.), (Berlin, Heidelberg), pp.~937--940, Springer Berlin Heidelberg, 2013.

\bibitem{WENTINK2014391}
E.~Wentink, V.~Schut, E.~Prinsen, J.~Rietman, and P.~Veltink, ``Detection of the onset of gait initiation using kinematic sensors and emg in transfemoral amputees,'' {\em Gait \& Posture}, vol.~39, no.~1, pp.~391--396, 2014.

\bibitem{mesin_crosstalk_2020}
L.~Mesin, ``Crosstalk in surface electromyogram: literature review and some insights,'' {\em Physical and Engineering Sciences in Medicine}, vol.~43, pp.~481--492, June 2020.

\bibitem{NAZARAHARI2022110880}
M.~Nazarahari, A.~Khandan, A.~Khan, and H.~Rouhani, ``Foot angular kinematics measured with inertial measurement units: A reliable criterion for real-time gait event detection,'' {\em Journal of Biomechanics}, vol.~130, p.~110880, 2022.

\bibitem{ballesteros_online_2016}
J.~Ballesteros, C.~Urdiales, A.~B. Martinez, and M.~Tirado, ``Online estimation of rollator user condition using spatiotemporal gait parameters,'' in {\em 2016 {IEEE}/{RSJ} {International} {Conference} on {Intelligent} {Robots} and {Systems} ({IROS})}, pp.~3180--3185, Oct. 2016.
\newblock ISSN: 2153-0866.

\bibitem{YOUDAS2005394}
J.~W. Youdas, B.~J. Kotajarvi, D.~J. Padgett, and K.~R. Kaufman, ``Partial weight-bearing gait using conventional assistive devices,'' {\em Archives of Physical Medicine and Rehabilitation}, vol.~86, no.~3, pp.~394--398, 2005.

\bibitem{ojeda_automatic_2018}
M.~Ojeda, A.~Cortés, J.~Béjar, and U.~Cortés, ``Automatic classification of gait patterns using a smart rollator and the {BOSS} model,'' {\em Proceedings of the 11th PErvasive Technologies Related to Assistive Environments Conference}, pp.~384--390, June 2018.
\newblock Conference Name: PETRA '18: The 11th PErvasive Technologies Related to Assistive Environments Conference ISBN: 9781450363907 Place: Corfu Greece Publisher: ACM.

\bibitem{MOSTAFAVI2022102173}
A.~Mostafavi, M.~R. Zakerzadeh, A.~Sadighi, and M.~A. Chalaki, ``An efficient design of an energy harvesting backpack for remote applications,'' {\em Sustainable Energy Technologies and Assessments}, vol.~52, p.~102173, 2022.

\bibitem{sierra_m_humanrobotenvironment_2019}
S.~D.~S. Sierra~M., M.~Garzón, M.~Múnera, and C.~A. Cifuentes, ``Human–{Robot}–{Environment} {Interaction} {Interface} for {Smart} {Walker} {Assisted} {Gait}: {AGoRA} {Walker},'' {\em Sensors}, vol.~19, p.~2897, June 2019.

\bibitem{alwan_basic_2007}
M.~Alwan, A.~Ledoux, G.~Wasson, P.~Sheth, and C.~Huang, ``Basic walker-assisted gait characteristics derived from forces and moments exerted on the walker's handles: {Results} on normal subjects,'' {\em Medical Engineering \& Physics}, vol.~29, pp.~380--389, Apr. 2007.

\bibitem{additive_stl}
G.~Dudek, ``Std: A seasonal-trend-dispersion decomposition of time series,'' {\em IEEE Transactions on Knowledge \& Data Engineering}, vol.~35, pp.~10339--10350, oct 2023.

\bibitem{soleymani2022surgical}
A.~Soleymani, X.~Li, and M.~Tavakoli, ``Surgical procedure understanding, evaluation, and interpretation: A dictionary factorization approach,'' {\em IEEE Transactions on Medical Robotics and Bionics}, vol.~4, no.~2, pp.~423--435, 2022.

\bibitem{bandara_mstl_2022}
K.~Bandara, R.~Hyndman, and C.~Bergmeir, ``{MSTL}: {A} {Seasonal}-{Trend} {Decomposition} {Algorithm} for {Time} {Series} with {Multiple} {Seasonal} {Patterns},'' {\em International Journal of Operational Research}, vol.~1, no.~1, p.~1, 2022.

\bibitem{STEIN2020931}
J.~Stein, ``Chapter 159 - stroke,'' in {\em Essentials of Physical Medicine and Rehabilitation (Fourth Edition)} (W.~R. Frontera, J.~K. Silver, and T.~D. Rizzo, eds.), pp.~931--936, Philadelphia: Elsevier, fourth edition~ed., 2020.

\bibitem{ballesteros_gait_2015}
J.~Ballesteros, C.~Urdiales, A.~B. Martinez, and M.~Tirado, ``Gait analysis for challenged users based on a rollator equipped with force sensors,'' in {\em 2015 {IEEE}/{RSJ} {International} {Conference} on {Intelligent} {Robots} and {Systems} ({IROS})}, pp.~5587--5592, Sept. 2015.

\bibitem{zifchock_symmetry_2008}
R.~A. Zifchock, I.~Davis, J.~Higginson, and T.~Royer, ``The symmetry angle: {A} novel, robust method of quantifying asymmetry,'' {\em Gait \& Posture}, vol.~27, pp.~622--627, May 2008.

\bibitem{soleymani2022domain}
A.~Soleymani, X.~Li, and M.~Tavakoli, ``A domain-adapted machine learning approach for visual evaluation and interpretation of robot-assisted surgery skills,'' {\em IEEE Robotics and Automation Letters}, vol.~7, no.~3, pp.~8202--8208, 2022.

\bibitem{soleymani2023artificial}
A.~Soleymani, X.~Li, and M.~Tavakoli, ``Artificial intelligence in robot-assisted surgery: Applications to surgical skills assessment and transfer,'' {\em Medical and Healthcare Robotics}, pp.~183--200, 2023.

\bibitem{9729350}
M.~A. Chalaki, D.~Maroufi, M.~Robati, M.~J. Karimi, and A.~Sadighi, ``An intelligent approach to detecting novel fault classes for centrifugal pumps based on deep cnns and unsupervised methods,'' in {\em 2021 7th International Conference on Signal Processing and Intelligent Systems (ICSPIS)}, pp.~01--06, 2021.

\end{thebibliography}

\end{document}